\pdfoutput=1

\documentclass[11pt]{article}

\usepackage[final]{acl}

\usepackage{times}
\usepackage{latexsym}

\usepackage{amsmath}
\usepackage{diagbox}
\usepackage{caption}
\usepackage{amsfonts,amssymb}
\usepackage{subfigure}
\usepackage{graphicx}
\usepackage{bm}
\usepackage{multirow}
\usepackage[ruled,lined,commentsnumbered]{algorithm2e}
\usepackage{hyperref}
\usepackage{booktabs}
\usepackage{bbm}
\usepackage[inkscapelatex=false]{svg}
\usepackage{pifont}
\usepackage{stfloats}
\usepackage{multicol}
\usepackage{float}
\usepackage{array}
\newcommand{\PreserveBackslash}[1]{\let\temp=\\#1\let\\=\temp}
\newcolumntype{C}[1]{>{\PreserveBackslash\centering}p{#1}}
\newcolumntype{R}[1]{>{\PreserveBackslash\raggedleft}p{#1}}
\newcolumntype{L}[1]{>{\PreserveBackslash\raggedright}p{#1}}

\usepackage{mathrsfs}

\usepackage[T1]{fontenc}

\usepackage[utf8]{inputenc}

\usepackage{microtype}

\usepackage{inconsolata}

\title{SiLLM: Large Language Models for Simultaneous Machine Translation}

\author{
    Shoutao Guo \textsuperscript{\rm 1,2},
    Shaolei Zhang \textsuperscript{\rm 1,2},
    Zhengrui Ma \textsuperscript{\rm 1,2},
    Min Zhang \textsuperscript{\rm 3},
    Yang Feng \textsuperscript{\rm 1,2}\thanks{ \ \ Corresponding author: Yang Feng.} \\
        \textsuperscript{\rm 1}{Key Laboratory of Intelligent Information Processing,} \\ Institute of Computing Technology, Chinese Academy of Sciences (ICT/CAS) \\
    { \textsuperscript{\rm 2} {University of Chinese Academy of Sciences, Beijing, China}} \\
    { \textsuperscript{\rm 3} {School of Future Science and Engineering, Soochow University}} \\
     \texttt{\{\href{mailto:guoshoutao22z@ict.ac.cn}{guoshoutao22z}, \href{mailto:zhangshaolei20z@ict.ac.cn}{zhangshaolei20z}, \href{mailto:fengyang@ict.ac.cn}{fengyang}\}@ict.ac.cn}  }

\begin{document}
\maketitle
\begin{abstract}



Simultaneous Machine Translation (SiMT) generates translations while reading the source sentence, necessitating a policy to determine the optimal timing for reading and generating words. Despite the remarkable performance achieved by Large Language Models (LLM) across various NLP tasks, existing SiMT methods predominantly focus on conventional transformers, employing a single model to concurrently determine the policy and generate the translations. However, given the complexity of SiMT, it is challenging to effectively address both tasks with a single model. Therefore, there is a need to decouple the SiMT task into policy-decision and translation sub-tasks. We propose SiLLM, which delegates the two sub-tasks to separate agents, thereby incorporating LLM into SiMT. The policy-decision agent is managed by a conventional SiMT model, responsible for determining the translation policy. The translation agent, leveraging the capabilities of LLM, generates translation using the partial source sentence. The two agents collaborate to accomplish SiMT. To facilitate the application of token-level policies determined by conventional SiMT models to LLM, we propose a word-level policy adapted for LLM. Experiments on two datasets demonstrate that, with a small amount of data for fine-tuning LLM, SiLLM attains state-of-the-art performance\footnote{Code is at \url{https://github.com/ictnlp/SiLLM}.}.
\end{abstract}

\section{Introduction}

Simultaneous Machine Translation (SiMT) \citep{DBLP:conf/acl/MaHXZLZZHLLWW19, DBLP:conf/iclr/MaPCPG20, ITST} starts generating translations before reading the entire source sentence, hence it is widely employed in scenarios with streaming input such as online international conferences and real-time subtitles. To obtain good trade-offs between latency and translation quality, SiMT models typically require a policy to determine when to read source words and when to generate target translation.

Current SiMT methods are predominantly based on traditional Transformer under the encoder-decoder framework \citep{DBLP:conf/nips/VaswaniSPUJGKP17}. According to the mechanism of translation policy, SiMT methods can be broadly categorized into fixed policy and adaptive policy. The SiMT models with fixed policy \citep{DBLP:conf/acl/MaHXZLZZHLLWW19, elbayad2020efficient, DBLP:conf/emnlp/ZhangF21} typically rely on external heuristic rules to perform translation when a fixed number of source words are read at each step, without considering the relationships between source sentence and target sentence. This may force the model to generate translation based on insufficient or redundant source information, resulting in degraded translation performance \citep{guo-etal-2023-learning}. In contrast, the SiMT methods with adaptive policy \citep{DBLP:conf/iclr/MaPCPG20, liu-etal-2021-cross} can dynamically control the moment to read or to translate by modeling the semantic correspondence between the input source words and generated translation, thereby achieving improved translation performance \citep{ITST}. 
Despite the difference in the mechanism of policy, existing SiMT methods need to introduce an additional module into the base model for policy-decision, together with the base model performing translation. However, this exceeds the capability of a single model.

Large Language Models (LLM) have achieved remarkable performance in multiple natural language processing tasks such as summarization and translation \citep{brown2020language, touvron2023llama}, and SiMT is expected to benefit from the strong abilities of LLM in understanding and generation, but unfortunately there is limited exploration of LLM for SiMT currently. This limitation can be attributed to the working mechanism of SiMT, which is challenging for the LLM to handle. SiMT requires a joint operation of policy-decision and translation while LLM can only manage translation generation in the absence of a policy-decision module. Additionally, SiMT has to process streaming input, but LLM cannot update their prompt inputs dynamically to fit the growth of the input.

On these grounds, we introduce SiLLM, a framework designed to conduct SiMT with LLM by employing a collaborative approach involving a policy-decision agent and a translation agent \citep{li2024agents}. The policy-decision agent employs a conventional SiMT model based on the encoder-decoder architecture \citep{DBLP:conf/nips/VaswaniSPUJGKP17} to make it compatible to utilize various existing SiMT models, while the translation agent adopts an LLM to take its advantage in understanding and generation where no restraints are added to the architecture of the LLM. In addition to the use of the two agents which can address the problem of lacking a policy module in LLM, SiLLM also maintains a memory to deal with streaming input, which is used to store both input source words and generated translation. 

All the components of SiLLM work collaboratively as follows. Based on the source words and generated words in memory, the policy-decision agent first determines whether to read source words to memory or to generate translation. Once the decision of generation is made, the translation agent will retrieve source words and translation in memory, and concatenate them as the prompt to proceed with the translation by predicting the next word. When the translation agent finishes generating the next target word, it will be added to the translation in memory. Then a new working cycle begins with the policy-decision agent. Additionally, conventional SiMT models all employ the token-level policies \citep{ DBLP:journals/corr/abs-2303-00257}, and applying them directly to LLM poses a vocabulary mismatch problem. We propose a novel word-level policy derived from token-level policy to address the problems.

The experiments demonstrate that, with a small amount of data fine-tuning LLM, our approach achieves significant improvement and attains state-of-the-art performance in SiMT.

\section{Background}
Our SiLLM utilizes the collaboration of the traditional transformer-based SiMT model and LLM to accomplish the SiMT task. Therefore, we provide a brief introduction to the SiMT task, transformer-based SiMT model, and LLM.

\paragraph{Simultaneous Machine Translation} The SiMT model incrementally reads the source sentence $\mathbf{x}$ = $(x_1, ..., x_J)$ with length $J$ and generates translation $\mathbf{y}$ = $(y_1, ..., y_I)$ with length $I$ based on the translation policy. To describe the translation policy, we denote $g_i$, which represents the number of input source words when generating the target word $y_i$. Then the policy for sentence pair ($\mathbf{x}, \mathbf{y}$) can be formulated as $\mathbf{g}$ = $(g_1, ..., g_I)$. During training, the SiMT model learns to generate target words using partial source words under the guidance of the policy $\mathbf{g}$ \citep{DBLP:conf/acl/MaHXZLZZHLLWW19}:
\begin{equation}
\mathcal{L}_{simt} = - \sum\limits_{i = 1}^{I} \log p(y_i \mid \mathbf{x}_{\leq g_i}, \mathbf{y}_{<i}, \theta_{simt}),
\end{equation}
where $\theta_{simt}$ represents the trained parameters of the SiMT model.

\paragraph{Hidden Markov Transformer} Built upon the Transformer architecture \citep{DBLP:conf/nips/VaswaniSPUJGKP17}, Hidden Markov Transformer (HMT) \citep{DBLP:journals/corr/abs-2303-00257} is the currently most advanced SiMT method. By incorporating the Hidden Markov Model, it models the translation policy as the hidden events and the target sentence as observed events. Consequently, the HMT model implicitly learns to employ multiple policies for translation during training through the formulation:
\begin{equation}
\mathcal{L}_{hmt} = - \log \sum\limits_{\mathbf{g}} p ( \mathbf{y}\mid \mathbf{x},\mathbf{g},\theta_{hmt} ) \times p( \mathbf{g}, \theta_{hmt}),
\end{equation}
where $\theta_{hmt}$ represents the trained parameters of HMT.
During inference, HMT strategically selects the most suitable policy to guide the model in generating translations. It is evident that HMT is responsible for both policy-decision and translation tasks. However, in our approach, we exclusively harness its capability for policy-decision.

\paragraph{Large Language Models} 
Large Language Models (LLM) commonly adopt the Decoder-only architecture \citep{brown2020language, touvron2023llama}. During training, LLM is trained on large corpora using the maximum likelihood estimation objective. As a result, LLM acquires a general capability to address various NLP tasks. For the translation task, it takes the concatenation of available source words $\mathbf{x}_{\leq j}$ and generated translation $\mathbf{y}_{<i}$ as input to generate the next target word $y_i$:
\begin{equation}
 p ( y_i\mid \mathbf{x}_{\leq j},\mathbf{y}_{<i},\theta_{LLM} ),
\end{equation}
where $\theta_{LLM}$ signifies the parameters of LLM. In our method, we leverage its translation capability.

\section{Method}
In this section, we introduce SiLLM, which leverages the collaboration of the policy-decision agent and the translation agent to jointly accomplish SiMT. We first outline the framework of SiLLM and its modules. To accomplish the vocabulary mismatch problem, we introduce a novel word-level policy, which derives from the token-level policy employed by conventional SiMT methods. Furthermore, we incorporate Supervised Fine-Tuning to enhance the translation capability of LLM. The details will be introduced in the following sections.

\subsection{Model Framework}
\begin{figure*}[t]
    \centering
    \includegraphics[width=6in]{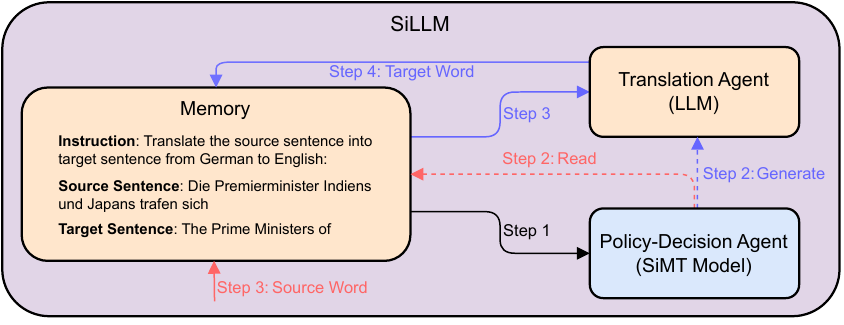}
    \caption{The framework of SiLLM. The numbers in the diagram signify the execution sequence of SiLLM. The red lines denote operations performed when the policy-decision agent determines to read source words. The blue lines indicate operations carried out when the decision for generation is made. The black line denotes the operation shared between both decision types.}
    \label{fig-model}
\end{figure*}

We introduce the model framework of SiLLM. As illustrated in Figure \ref{fig-model}, it primarily consists of three modules: memory, policy-decision agent, and translation agent. The memory module stores the instructions, input source words, and generated translation. The policy-decision agent is managed by the conventional SiMT model \cite{DBLP:conf/iclr/MaPCPG20, DBLP:journals/corr/abs-2303-00257}, which utilizes existing source words and generated target words to determine whether to read or generate a word. The translation agent employs an LLM to generate translation when the decision to generate is made.

At each step, the policy-decision agent retrieves existing source words and generated target words from memory to determine the action. When the policy-decision agent decides to read source words, SiLLM will read a source word and store it in memory. When the decision to generate target words is made, the policy-decision agent activates the translation agent. The translation agent retrieves instructions, existing source words, and generated translations from the memory module and then uses this information to generate the next target word. Subsequently, the translation agent places the newly generated word into the memory module. Throughout the entire process, the two agents collaborate to accomplish the SiMT task.

In SiLLM, any conventional SiMT model can serve as the policy-decision agent, and any LLM can be employed as the translation agent. In our setup, we use \texttt{Llama2-7B-chat} \citep{touvron2023llama} as translation agent and HMT model \citep{DBLP:journals/corr/abs-2303-00257} as the policy-decision agent.

\subsection{Word-level Policy}
After introducing the framework, we propose a novel word-level translation policy. The conventional SiMT models \citep{DBLP:conf/iclr/MaPCPG20, DBLP:journals/corr/abs-2303-00257} yield a token-level policy. However, employing the token-level policy directly for translation poses challenges for LLM, due to vocabulary mismatch. Therefore, we introduce the word-level policy, derived from the token-level policy. Furthermore, to ensure that the performance is not affected by outlier policies, we incorporate boundary restrictions for the word-level policy. The methods for acquiring the word-level policy and implementing boundary restrictions are introduced sequentially.

\paragraph{Word-level Policy} Given the source sentence $\mathbf{x}$ = $(x_1, ..., x_J)$ and the target sentence $\mathbf{y}$ = $(y_1, ..., y_I)$, we define the corresponding source token sequence and target token sequence as $\mathbf{s}$ = $(s_1, ..., s_M)$ and $\mathbf{t}$ = $(t_1, ..., t_N)$, respectively. By employing the conventional SiMT model, we can obtain the token-level policy $\mathbf{h}$ = $(h_1, ..., h_N)$, where $h_n$ denotes the number of available source tokens when translating $t_n$. Subsequently, we introduce the method of transforming token-level policy into the word-level policy $\mathbf{g}$ = $(g_1, ..., g_I)$, where $g_i$ is the number of source words when translating $y_i$.

For the target token sequence $\mathbf{t}$, we iterate through each target token sequentially. When the target token $t_n$ is detected as the last token of a target word $y_i$, we obtain the number of source tokens $h_n$ used for translation at that moment. 
By assessing the number of complete source words formed by the first $h_n$ source tokens, we determine the number of complete source words as $u$ when translating $y_i$. Then $g_i$ is calculated as:
\begin{equation}
    g_i = \min \{u+1, J\},
\end{equation}
where $J$ is the length of the source sentence.
After traversing the entire target token sequence $\mathbf{t}$, we can obtain the word-level policy $\mathbf{g}$.

\begin{figure}[t]
    \centering
    \includegraphics[width=3in]{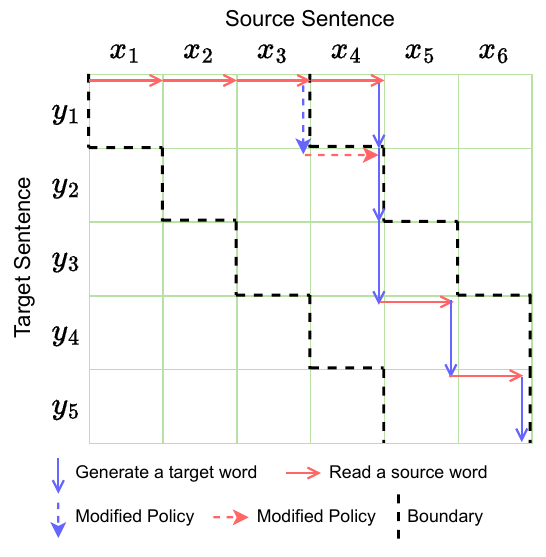}
    \caption{The illustration of incorporating boundary restrictions to word-level policy. The hyperparameters $B$ and $T$ in the figure are set to 1 and 3, respectively. In the absence of boundary restrictions, the word-level policy generates $y_1$ after reading $x_4$. However, our approach modifies it to generate $y_1$ upon reading $x_3$. }
    \label{boundary_constrain}
\end{figure}

\paragraph{Boundary Restrictions}

Relying solely on the conventional SiMT model to determine the policy sometimes proves inadequate in preventing the emergence of outlier policies \citep{DBLP:journals/corr/abs-2303-00257}. These outlier policies may lead the model to generate translations by exploiting insufficient source information or result in excessively high latency \citep{DBLP:conf/iclr/MaPCPG20}. To address the problem of outlier policies, we impose restrictions on the obtained word-level policy $\mathbf{g}$ = $(g_1, ..., g_I)$. We then elaborate on the method of adding boundaries.

We define hyperparameters $B$ and $T$ to represent the minimum and maximum number of source words that can be considered when translating the first target word, respectively. The word-level policy with boundary restrictions can be obtained using Eq.(\ref{eq_1}) and Eq.(\ref{eq_2}) sequentially:
\begin{equation}
\label{eq_1}
    g^r_i = \min \{\max \{g_i, i-1+B\}, i-1+T\},
\end{equation}
\begin{equation}
\label{eq_2}
    g^r_i = \min \{g^r_i, J\},
\end{equation}
where $J$ is the length of the source sentence. Therefore, we obtain the word-level policy $\mathbf{g}^r$ = $(g^r_1, ..., g^r_I)$ with boundary restrictions. Figure \ref{boundary_constrain} provides an intuitive illustration of the word-level policy with boundary restrictions.

\subsection{Supervised Fine-Tuning}

After introducing the word-level policy, the various modules of SiLLM can collaborate to effectively carry out the SiMT task. To further improve the overall performance of SiLLM, we fine-tune the LLM using Supervised Fine-Tuning (SFT), thereby enhancing the translation capabilities of LLM acting as the translation agent. As the translation agent, LLM typically utilizes partial source information to generate translations. Previous studies \citep{DBLP:conf/iclr/MaPCPG20, liu-etal-2021-cross, ITST} highlights that the SiMT model can demonstrate effective performance during inference only if it is trained to learn the ability to translate based on partial source sentence. However, there is no dedicated parallel corpus for the SiMT task, where the input is the source prefix and the output is the corresponding target prefix.

In our method, we fine-tune the LLM using a limited amount of full-sentence parallel corpus. Notably, despite not being explicitly trained on SiMT corpora, our method exhibits a significant improvement over the SiMT model built upon the Transformer architecture \citep{DBLP:conf/nips/VaswaniSPUJGKP17}, achieving state-of-the-art performance. This underscores that our approach can effectively stimulate the LLM to excel in generating translations based on partial source information. 

During SFT, SiLLM fine-tunes LLM with LoRA \citep{hu2021lora}, and the details are introduced in the Section \ref{Experiments}.

\section{Experiments}
\label{Experiments}
\subsection{Datasets}
We primarily validate our approach on two translation tasks, each representing high-resource and low-resource translation scenarios. We will provide a brief introduction to each dataset.

\textbf{WMT15\footnote{\url{www.statmt.org/wmt15}} German$\rightarrow$English (De$\rightarrow$En)} This dataset comprises 4.5M parallel sentence pairs. During SFT, we randomly sample 100k sentence pairs for fine-tuning LLM. Consistent with \citet{DBLP:conf/iclr/MaPCPG20}, we use newstest2013 as the validation set and newstest2015 as the test set.

\textbf{MuST-C English$\rightarrow$German (En$\rightarrow$De)} We conduct translation on text data \citep{di-gangi-etal-2019-must}. This dataset contains 230k samples. Similarly, we randomly sample 100k samples from the training set for fine-tuning LLM. We use dev set for validation and tst-COMMON set for the testing.

\begin{figure*}[t]
\centering
\subfigure[Comparison of SiLLM and previous Transformer-based SiMT methods on De$\rightarrow$En task.]{
\includegraphics[width=1.9in]{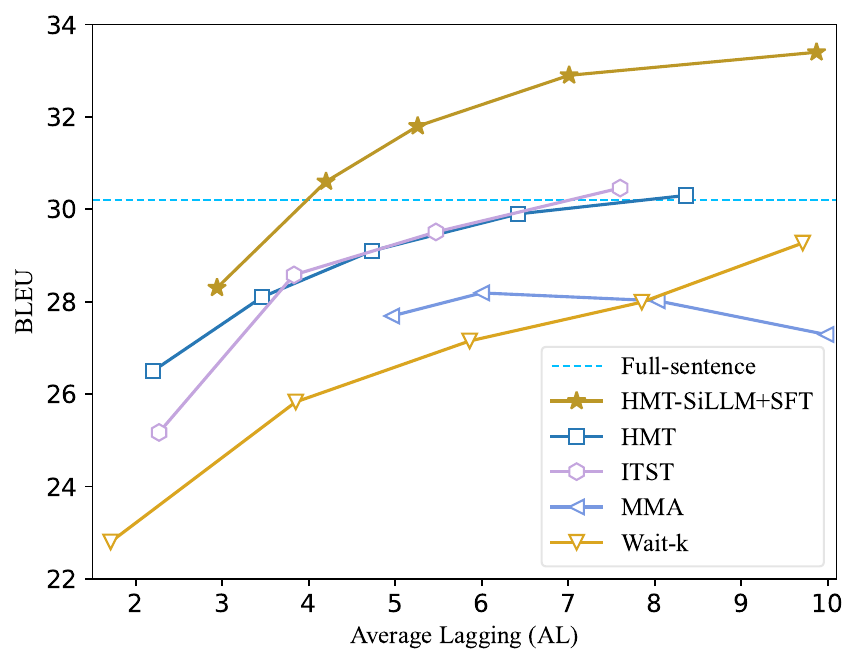}
}\hspace{0.1cm}
\subfigure[The comparison of HMT with the variants of SiLLM on De$\rightarrow$En task.]{
\includegraphics[width=1.99in]{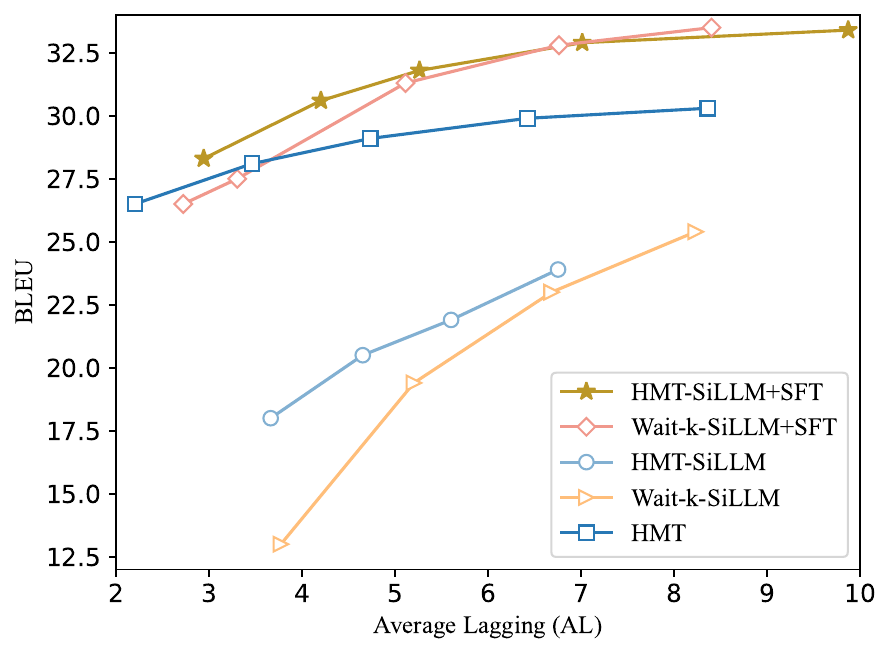}
}\hspace{0.1cm}
\subfigure[The comparison of HMT with the variants of SiLLM on En$\rightarrow$De task.]{
\includegraphics[width=1.9in]{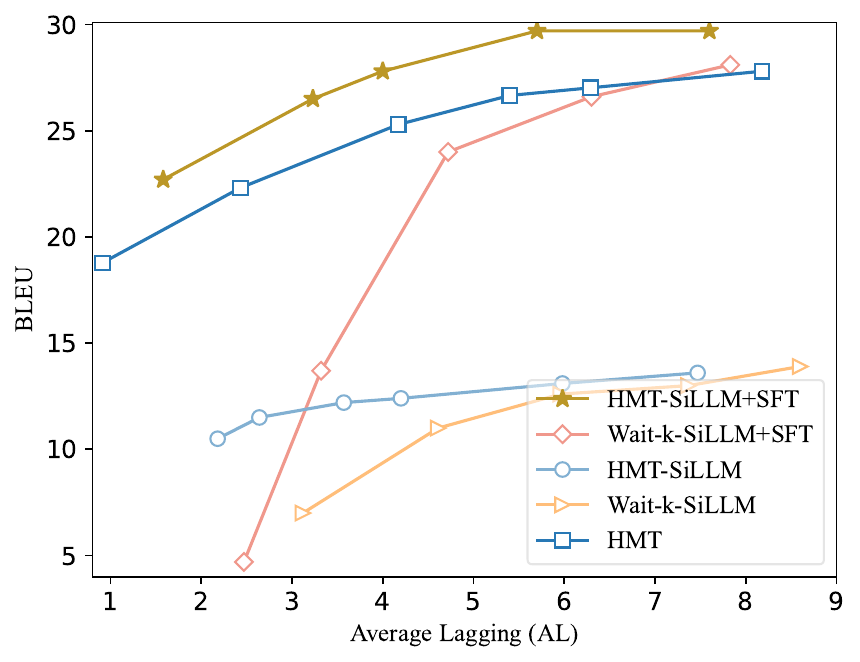}
}

\caption{Performance of different SiMT methods on De$\rightarrow$En and En$\rightarrow$De tasks.}
\label{main_res}
\end{figure*}

\subsection{System Settings}
Our experiments involve the following methods. Apart from the methods based on SiLLM, other SiMT methods employ Transformer architecture \citep{DBLP:conf/nips/VaswaniSPUJGKP17}. The methods based on SiLLM use \texttt{Llama2-7B-chat}\footnote{\url{https://huggingface.co/meta-llama/Llama-2-7b-chat-hf}} as translation agent.

\textbf{Full-sentence} is the conventional full-sentence machine translation model, which adopts the encoder-decoder architecture.

\textbf{Wait}-$k$ initially reads $k$ source tokens, and then generates a target token and reads a source token alternately \citep{DBLP:conf/acl/MaHXZLZZHLLWW19}.

\textbf{MMA} is the first Transformer-based SiMT model that performs the adaptive policy. It allows each head to independently determine its policy and integrates the outputs of multiple heads to generate translations \citep{DBLP:conf/iclr/MaPCPG20}.

\textbf{ITST} models the SiMT task of translating source sentence to target sentence as an optimal transport problem and determines the policy by accumulating source information \citep{ITST}.

\textbf{HMT} is the current state-of-the-art adaptive policy, utilizing a Hidden Markov Model to model SiMT task \citep{DBLP:journals/corr/abs-2303-00257}.

\textbf{Wait-$k$-SiLLM} adopts the Wait-$k$ as the policy-decision agent without fine-tuning LLM.

\textbf{HMT-SiLLM} utilizes the HMT model as the policy-decision agent without fine-tuning LLM.

\textbf{Wait-$k$-SiLLM+SFT} further fine-tunes LLM upon Wait-$k$-SiLLM.

\textbf{HMT-SiLLM+SFT} fine-tunes LLM based on HMT-SiLLM.

The SiMT methods based on Transformer are all adapted from Fairseq Library \citep{DBLP:conf/naacl/OttEBFGNGA19}. They apply Transformer-Base (6 layers, 8 heads) for the De-En task and Transformer-Small (6 layers, 4 heads) for the En-De task.

The methods based on SiLLM are all implemented using the Alpaca-LoRA Library\footnote{\url{https://github.com/tloen/alpaca-lora}}. During SFT, we fine-tune the LLM using LoRA \citep{hu2021lora}. For the adapters of LoRA, $r$ is set to 8, and $\alpha$ is set to 16. We set the learning rate to 0.0001 and batch size to 128. More training details are shown in Appendix \ref{exper_detail}.

During inference, we apply greedy search to all methods. The methods based on SiLLM adopt the word-level policy. We evaluate all methods with latency measured by Average Lagging (AL) \citep{DBLP:conf/acl/MaHXZLZZHLLWW19} and translation quality estimated by SacreBLEU \citep{post-2018-call}.

\subsection{Main Results}
We compare SiLLM with previous SiMT methods built upon Transformer and explore different settings of SiLLM. As shown in Figure \ref{main_res}, HMT-SiLLM+SFT exhibits a significant improvement and achieves state-of-the-art performance.

HMT-SiLLM+SFT demonstrates a significant performance improvement over previous SiMT methods. Previous SiMT methods necessitate a single model to concurrently manage policy-decision and translation \citep{DBLP:conf/iclr/MaPCPG20}. This exceeds the capacity of Transformer-based SiMT models, leading to degraded performance. In contrast, our approach decomposes the SiMT task into policy-decision and translation sub-tasks, assigning them to separate agents. Each agent is responsible for a specific sub-task and collaborates with another agent to complete the SiMT, fully leveraging the strengths of each agent. Additionally, due to limitations in source information, the gap between HMT-SiLLM+SFT and traditional SiMT methods is relatively smaller at low latency. However, as latency increases, the divergence in translation capabilities becomes more pronounced.

\begin{table}[]
\centering
\begin{tabular}{c c c  c} \toprule[1.2pt]
\textbf{SFT Data} & $k$ & \textbf{AL} & \textbf{BLEU}                                       \\ \cmidrule(lr){1-1} \cmidrule(lr){2-2}\cmidrule(lr){3-4}

Full-Sentence & 5 & \textbf{5.11} & \textbf{31.30} \\

SiMT & 5 & 5.43 &  1.00 \\

 \bottomrule[1pt]
\end{tabular}
\caption{The performance of Wait-$k$-SiLLM+SFT when using different training data for SFT. `Full-Sentence' denotes full-sentence pairs. `SiMT' represents the SiMT data corresponding to the Wait-5 policy.
The experiments are performed on the De$\rightarrow$En task. }
\label{data_type}
\end{table}

HMT-SiLLM+SFT attains the best performance among all variants of SiLLM. Compared to the Wait-$k$-SiLLM+SFT, HMT-SiLLM+SFT excels in policy-decision. HMT possesses the capability to dynamically adjust its policy \citep{DBLP:journals/corr/abs-2303-00257}, offering a more superior policy than the fixed policy of Wait-$k$ \citep{DBLP:conf/acl/MaHXZLZZHLLWW19}. This dynamic adjustment assists the translation agent in making well-informed decisions regarding translation timing, thereby achieving better performance. In comparison to HMT-SiLLM, the translation agent of HMT-SiLLM+SFT undergoes fine-tuning, enhancing its translation capabilities. Consequently, enhancing the capabilities of each agent in SiLLM contributes to an overall improvement in the performance of SiMT.

\section{Analysis}
To enhance the comprehension of SiLLM, we have undertaken various analyses, with all analytical experiments conducted on the De$\rightarrow$En task.

\subsection{Ablation Study}
\label{ablation}
In the ablation experiments, we primarily investigate the influence of training data used for SFT on the performance of SiMT. Since SFT exclusively affects the translation agent, it essentially explores how the training data impacts the translation performance of the LLM.

Previous SiMT methods \citep{DBLP:conf/iclr/MaPCPG20, ITST} have found that SiMT models excel in SiMT during inference only when trained to generate translations based on partial source sentences. To validate this in SiLLM, we construct the SiMT data corresponding to the Wait-5 policy to fine-tune the LLM and compare it with our method. As depicted in Table \ref{data_type}, full-sentence data proves more effective in stimulating the ability of LLM to translate based on incomplete source information compared to SiMT data corresponding to the Wait-$k$ policy. We find that the translations generated by LLM deviate from the source sentences and lack fluency after using SiMT data for SFT. We attribute this issue to the fact that the SiMT data corresponding to the Wait-$k$ policy fails to consider the semantic equivalence between the source and target, resulting in the failure of SFT. The construction method for the SiMT data is detailed in the Appendix \ref{construct_method}.

\begin{figure}[t]
    \centering
    \includegraphics[width=2.9in]{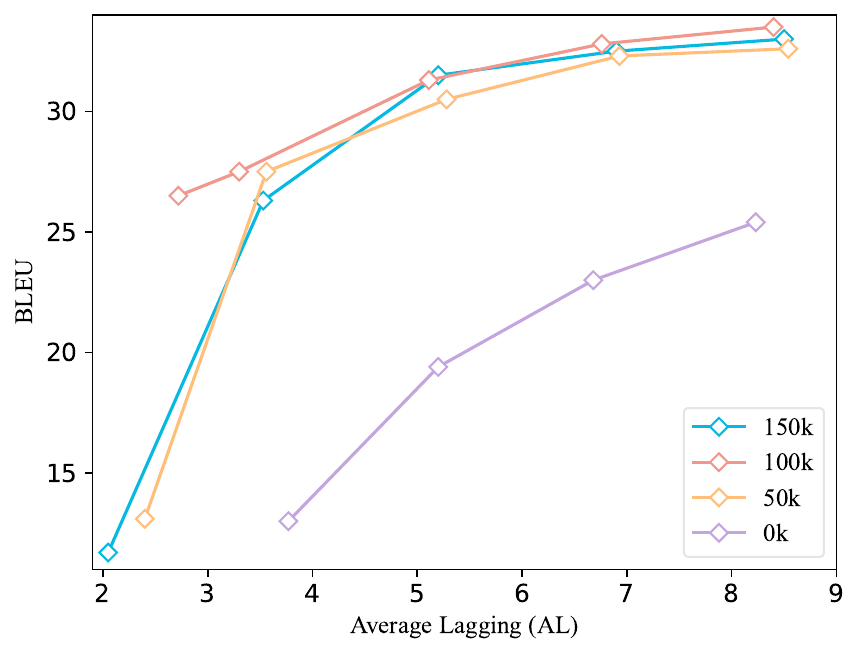}
    \caption{The impact of different quantities of training data during SFT on Wait-$k$-SiLLM+SFT. The experiments are conducted on the De$\rightarrow$En task}
    \label{SFT_data_items}
\end{figure}

Opting for fine-tuning LLM with full-sentence data, we further explore the impact of data quantity. Figure \ref{SFT_data_items} illustrates that the performance of SiLLM is significantly enhanced after SFT. Additionally, the impact of data quantity on SiMT performance appears to be relatively minimal. Therefore, we choose to fine-tune LLM with 100k data, which demonstrates relatively better performance under all latency.

\subsection{Hallucination Rate}
In SiMT, hallucination typically refers to instances where the words in translation lack semantic counterparts in the source sentence. There are generally two main reasons for hallucinations in SiMT. On one hand, it stems from the insufficient translation capability of the SiMT model, resulting in the generation of unrelated words \citep{zhang2020futureguided}. On the other hand, SiMT models are compelled to utilize unrelated source information to predict target words under the guidance of a policy \citep{ma-etal-2023-non}. Therefore, the hallucinations in translation reflect a combination of translation and policy-decision capabilities.

\begin{figure}[t]
    \centering
    \includegraphics[width=2.9in]{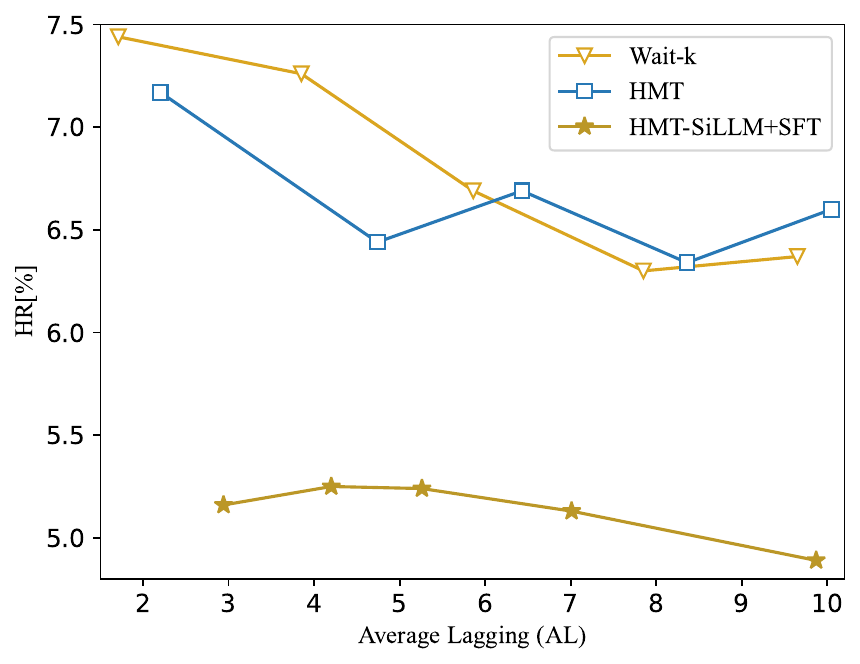}
    \caption{The hallucination rate (HR) of different SiMT methods. The results are based on the De$\rightarrow$En task.}
    \label{hall}
\end{figure}

\begin{table}[]
\centering
\begin{tabular}{c c c  c} \toprule[1.2pt]
\textbf{Method} & \textbf{AL} & \textbf{BLEU} & \textbf{Speed} $(\uparrow)$                                       \\ \cmidrule(lr){1-1} \cmidrule(lr){2-3}\cmidrule(lr){4-4}

Wait-$k$ & 3.85 & 25.83 & \textbf{107.97} \\

HMT & 4.73 &  29.10 & 32.47 \\

HMT-SiLLM+SFT & \textbf{4.2} & \textbf{30.60} & 9.94 \\

 \bottomrule[1pt]
\end{tabular}
\caption{Comparison of inference speed among different SiMT methods. `Speed' is measured by the number of words generated per second (i.e., word/s). The experiments are performed on the De$\rightarrow$En task.}
\label{inference_speed}
\end{table}

To quantify the hallucination in translation, we introduce the hallucination rate (HR) \citep{DBLP:conf/emnlp/ChenZKM021}, which measures the proportion of hallucinated words in the translation. We employ \texttt{eflomal}\footnote{\url{https://github.com/robertostling/eflomal}} to get the alignments between the source sentence and translation, thereby calculating the HR for various SiMT methods. Figure \ref{hall} demonstrates that our method exhibits lower hallucination in translation under all latency. Compared to Wait-$k$, HMT achieves a lower HR due to its ability to adjust policies based on the translation status. Our approach decomposes the SiMT task into two sub-tasks and assigns them to different agents, preventing overburdening a single model. This maximizes the benefits of both policy-decision and translation, thereby achieving better trade-offs between latency and translation quality.

\begin{figure}[t]
    \centering
    \includegraphics[width=2.9in]{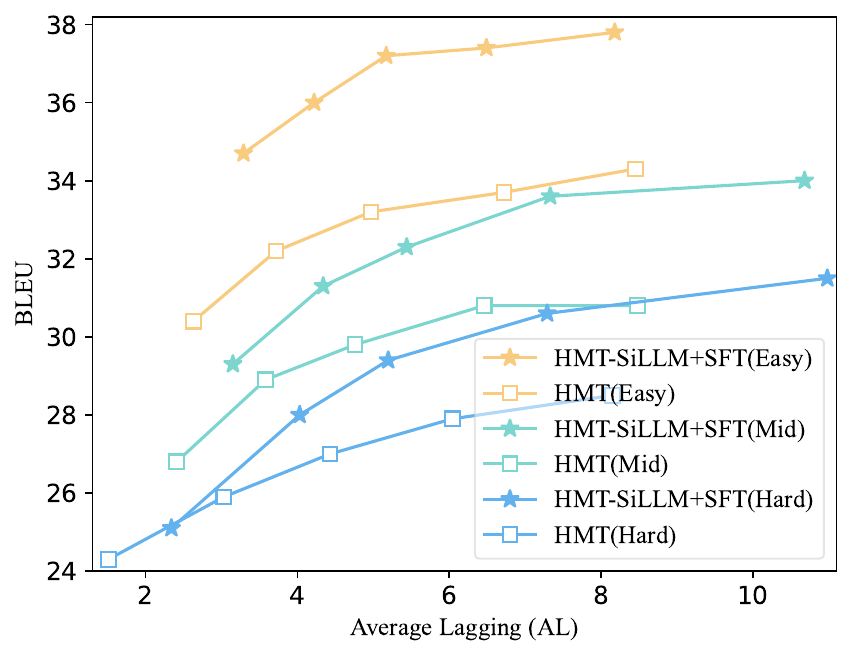}
    \caption{Translation performance of SiMT methods at different levels of difficulty. The experiments are conducted on the De$\rightarrow$En dataset.}
    \label{difficult}
\end{figure}
\subsection{Inference Speed}

The evaluation of SiMT tasks involves the assessment of latency and translation quality. However, the latency metric does not reflect the inference speed of machine \citep{DBLP:conf/acl/MaHXZLZZHLLWW19}. When incorporating LLM into SiMT methods, a crucial question arises regarding whether the inference speed of LLM could potentially act as an application bottleneck for the SiMT model. Therefore, it is important to compare the actual inference speed of our method with that of previous SiMT methods to demonstrate the practical usability of our approach.

We evaluate the inference speed of various SiMT methods using NVIDIA GeForce RTX 3090 and present the results in Table \ref{inference_speed}. The Wait-$k$ method exhibits the poorest performance but attains the highest inference speed due to its simplistic architecture. Although slower than the Wait-$k$, HMT demonstrates a significant improvement in performance. Our approach, incorporating HMT and LLM, achieves a slower inference speed but optimal performance. Considering the application scenarios of the SiMT task, our method can generate 9 words per second, fully meeting practical requirements. Moreover, our method can benefit from acceleration on more advanced hardware, resulting in faster inference speeds.

\subsection{Performance at Different Levels of Difficulty}

The word reordering (i.e., non-monotonic alignment) between the source and target sentences presents challenges for policy-decision and translation capabilities in SiMT. The translation difficulty varies across different source sentences, with long-distance non-monotonic alignments making translation notably more challenging. Consequently, we investigate the performance of SiMT models at different levels of translation difficulty. Based on the number and distance of non-monotonic alignment between the source sentence and ground-truth, we evenly divide the test set for the De$\rightarrow$En task into three levels: Easy, Medium, and Hard. Following the division, the Easy set primarily consists of monotonic alignments, while the Hard set includes sentences with at least 12 non-monotonic alignments \citep{DBLP:conf/emnlp/ZhangF21}.

In Figure \ref{difficult}, we present the evaluation results of our method and HMT on datasets of three difficulty levels. Our method demonstrates substantial improvements over HMT at both easy and medium levels across all latency. However, at the difficult level, the performance of our method is relatively close to HMT at low latency. This is due to the extensive word reordering, which restricts the accessibility of source information for SiMT models at low latency. As the latency increases, the gap between our method and HMT also widens, underscoring the importance of task decomposition in our approach to prevent excessive burden on a single model.

\section{Related Work}

Simultaneous Machine Translation (SiMT) generates translations while reading the source sentence \citep{reinforcement}. This necessitates a translation policy to determine the number of source words when translating each target word. Consequently, SiMT entails the combination of policy-decision and translation. Depending on whether the policy is determined by the SiMT model, SiMT methods can be categorized into fixed and adaptive policies.

For the fixed policy, the SiMT model is guided by external heuristic rules to generate translations. \citet{DBLP:conf/acl/MaHXZLZZHLLWW19} proposes the Wait-$k$ policy and introduces a more reasonable latency metric called average lagging. \citet{elbayad2020efficient} introduces the multi-path training method and first utilizes a unidirectional encoder to address the issue of re-encoding. Additionally, \citet{zhang2020futureguided} enhances the prediction capabilities of SiMT models by minimizing the distance between the bidirectional and unidirectional representations of the source sentences. \citet{DBLP:conf/emnlp/ZhangF21} proposes MoE Wait-$k$, which assigns a fixed policy to each head and generates translation by integrating the results of all heads.

For adaptive policy, the SiMT model determines the policy and generates translation concurrently. \citet{DBLP:conf/iclr/MaPCPG20} proposes MMA, which allows each head to independently determine the policy and makes all heads jointly decide on the translation. \citet{DBLP:conf/emnlp/MiaoBS21} proposes a generative SiMT framework, which uses a re-parameterized Poisson prior to regularize the policy. \citet{ITST} models the SiMT task as an optimal transport problem and determines the policy by assessing the sufficiency of source information. \citet{DBLP:journals/corr/abs-2303-00257} proposes HMT, which utilizes the Hidden Markov Model to handle the SiMT task and achieves state-of-the-art performance. However, all these methods involve a single model responsible for both policy decision and translation, surpassing the capability of a single model.

We propose to decompose the SiMT task into policy-decision and translation sub-tasks and assign them to different agents. The translation agent is managed by LLM, while the policy-decision agent is undertaken by a Transformer-based SiMT model. These agents collaborate to accomplish SiMT, fully utilizing the translation capability of the LLM and the policy-decision capability of the conventional SiMT model. This allows our method to achieve optimal performance.

\section{Conclusion}
In this paper, we introduce SiLLM, a novel framework that harnesses existing Transformer-based SiMT methods and LLM to collaboratively achieve SiMT. Experiments showcase that our method attains state-of-the-art performance and is practical for applications.

\section*{Limitations}
Throughout the experiments, we investigate the utilization of HMT as the policy-decision agent and \texttt{Llama2-7B-chat} as the translation agent, significantly surpassing the performance of previous SiMT methods. We believe that exploring more powerful translation agents or better policy-decision agents will contribute to further performance improvement.

\bibliography{custom}

\appendix

\section{Experiment Details}
\label{exper_detail}

\begin{figure*}[t]
    \centering
    \includegraphics[width=6in]{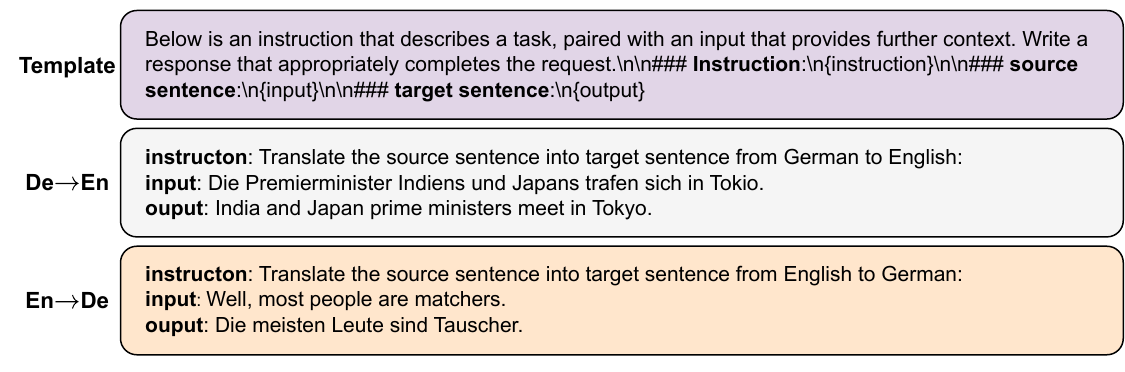}
    \caption{The prompt template for the LLM, along with examples for both De$\rightarrow$En and En$\rightarrow$De tasks.}
    \label{Instruct}
\end{figure*}

In this section, we primarily present the experimental setup for HMT-SiLLM+SFT. The settings of HMT-SiLLM+SFT are shown in Table \ref{tab:hyperparameter}. As shown in Figure \ref{Instruct}, we also provide the prompt template for the LLM, along with examples for both De$\rightarrow$En and En$\rightarrow$De tasks. 

\section{Methods of Constructing SiMT Data}
\label{construct_method}
In subsection \ref{ablation}, we fine-tune the LLM using SiMT data corresponding to the Wait-5 policy. The methodology for constructing the SiMT data is elucidated in this section.

For a given sentence pair ($\mathbf{x}$, $\mathbf{y}$), with $\mathbf{x}$ having a length of $J$ and $\mathbf{y}$ having a length of $I$, we construct SiMT data corresponding to the Wait-$k$ policy, denoted as prefix pair. If the condition $k \geq J$ is satisfied, the prefix pair is ($\mathbf{x}$, $\mathbf{y}$). Otherwise, we randomly select $j$ from the interval $[k, J]$, where $J$ is the length of the source sentence. In this scenario, the source prefix is $\mathbf{x}_{\leq j}$. The length of the corresponding target prefix is calculated as:
\begin{equation}
    i = \min(j-k+1, I),
\end{equation}
where $I$ is the length of the target sentence. Consequently, the corresponding target prefix becomes $\mathbf{y}_{\leq i}$. At this point, we acquire the prefix pair ($\mathbf{x}_{\leq j}$, $\mathbf{y}_{\leq i}$).

\section{Numeric Results of Main Experiment}
In addition to the results shown in the Figure \ref{main_res}, we also provide the numeric results of the main experiment for better comparison. Table \ref{con_deen} shows the results of conventional representative simultaneous SiMT methods on WMT15 De->En task. Table \ref{our_deen} demonstrates the results of SiLLM on the WMT15 De-En task. Table \ref{our_ende} gives the results of HMT and SiLLM on the MuST-C En->De task.

\begin{table*}[t]
\centering
\caption{Settings of HMT-SiLLM+SFT.}

\begin{tabular}{c|c|c|cc}
\toprule
\multicolumn{3}{c|}{\textbf{Hyperparameters}} & \textbf{WMT15 De$\rightarrow$En} & \textbf{MuST-C En$\rightarrow$De} \\
\bottomrule
\multirow{9}{*}{LLM} & Base\_model & Base\_model & \multicolumn{2}{c}{\texttt{Llama2-7B-chat}} \\
                    \cmidrule{2-5}
                     & \multirow{4}{*}{LoRA}  & lora\_r     & 8      & 8 \\
                     &                        & lora\_alpha & 16     & 16 \\
                     &                        & lora\_dropout & 0.05 & 0.05 \\
                     &                        & lora\_target\_modules & \multicolumn{2}{c}{q\_proj, k\_proj, v\_proj, o\_proj} \\
                     \cmidrule{2-5}
                     & \multirow{3}{*}{Training Details} & batch\_size & 128 & 128 \\
                     &                           & micro\_batch\_size & 4 & 4 \\
                     &                           & learning\_rate & 1e-4 & 1e-4 \\
\midrule
\multirow{19}{*}{HMT}  & \multirow{4}{*}{Encoder} & encoder\_layers & 6 & 6 \\
                     &                         & encoder\_embed\_dim & 512 & 512 \\
                    &                         & encoder\_ffn\_embed\_dim & 2048 & 1024 \\
                    &                           & encoder\_attention\_heads & 8 & 4 \\ 
                    \cmidrule{2-5}
                    & \multirow{4}{*}{Decoder} & decoder\_layers & 6 & 6 \\
                     &                         & decoder\_embed\_dim & 512 & 512 \\
                    &                         & decoder\_ffn\_embed\_dim & 2048 & 1024 \\
                    &                           & decoder\_attention\_heads & 8 & 4 \\
                    \cmidrule{2-5}
                    & \multirow{11}{*}{Training Details} & dropout   & 0.3    &0.3  \\
                    &                            &optimizer & adam  &adam   \\
                    &   &adam\_$\beta$          & (0.9, 0.98)   & (0.9, 0.98)     \\
                    & &clip\_norm        & 0             &0                \\
                    &&lr & 5e-4          & 5e-4                \\
                && lr\_scheduler   & inverse\_sqrt  & inverse\_sqrt     \\
                &&warmup\_updates       & 4000          & 4000            \\
&&warmup\_init\_lr     & 1e-7          & 1e-7                \\
&& weight\_decay       & 0.0        & 0.0        \\
&& label\_smoothing    & 0.1           & 0.1                \\
&&max\_tokens          &  8192$\times$4 & 8192$\times$4 \\
\bottomrule

\end{tabular}
\label{tab:hyperparameter}
\end{table*}

\begin{table}[]
\centering
\begin{tabular}{p{2cm}<{\centering} p{2cm}<{\centering} p{2cm}<{\centering}} 
\toprule[1.5pt]
\multicolumn{3}{c}{\textit{\textbf{Full-sentence}}}     \\\hline
          & AL  & SacreBLEU   \\
          & n/a    &30.21   \\
\midrule[1pt]
\multicolumn{3}{c}{\textit{\textbf{Wait-$k$}}}     \\
\hline
    $k$      & AL  & SacreBLEU   \\
    3      &1.71      &22.80  \\
    5      &3.85      &25.83  \\
    7      &5.86      &27.15  \\
    9      &7.85      &27.99  \\
    11      &9.71      &29.27  \\
\midrule[1pt]
\multicolumn{3}{c}{\textit{\textbf{MMA}}}     \\
\hline
    $\lambda$      & AL  & SacreBLEU   \\
    0.40     &4.96     &25.82   \\
    0.30      &6.00     &27.99  \\
    0.25    &8.03     &28.33  \\
    0.20   &9.98     &28.39  \\
\midrule[1pt]

\multicolumn{3}{c}{\textit{\textbf{ITST}}}     \\
\hline
    $\delta$    & AL  & SacreBLEU   \\
    0.20      &2.27       &25.17   \\
    0.40    &3.83  & 28.58  \\
    0.50      &5.47       &29.51   \\
    0.60    &7.60  & 30.46   \\
\midrule[1pt]

\multicolumn{3}{c}{\textit{\textbf{HMT}}}     \\
\hline
    $L, K$      & AL  & SacreBLEU   \\
    2, 4      &2.20      &26.50   \\
    3, 6     &3.46      &28.10  \\
    5, 6      &4.73      &29.10  \\
    7, 6     &6.42      &29.90  \\
    9, 8     &8.36      &30.30 \\

\bottomrule[1.5pt]

\end{tabular}
\caption{Numerical results of conventional SiMT methods on De$\rightarrow$En task.}
\label{con_deen}
\end{table}

\begin{table}[]
\centering
\begin{tabular}{p{2cm}<{\centering} p{2cm}<{\centering} p{2cm}<{\centering}} 
\toprule[1.5pt]

\multicolumn{3}{c}{\textit{\textbf{Wait-$k$-SiLLM}}}     \\
\hline
    $k$      & AL  & SacreBLEU   \\
    1      &3.77      &13.00  \\
    3      &5.20      &19.40  \\
    5      &6.68      &23.00  \\
    7      &8.23      &25.40  \\
\midrule[1pt]
\multicolumn{3}{c}{\textit{\textbf{Wait-$k$-SiLLM+SFT}}}     \\
\hline
    $k$      & AL  & SacreBLEU   \\
    2      &2.72      &26.50  \\
    3      &3.30      &27.50  \\
    5      &5.11      &31.30  \\
    7      &6.76      &32.80  \\
    9      &8.40      &33.50  \\
\midrule[1pt]

\multicolumn{3}{c}{\textit{\textbf{HMT-SiLLM}}}     \\
\hline
    $L, K$    & AL  & SacreBLEU   \\
    2, 4   &2.20      &26.50   \\
    3, 6   &3.46      &28.10  \\
    5, 6    &4.73      &29.10  \\
    7, 6  &6.42      &29.90  \\
\midrule[1pt]

\multicolumn{3}{c}{\textit{\textbf{HMT-SiLLM+SFT}}}     \\
\hline
    $L, K$    & AL  & SacreBLEU   \\
    3, 6   &2.94      &28.30  \\
    5, 6    &4.20      &30.60  \\
    7, 6   &5.26      &31.80  \\
    9, 8   &7.01      &32.90   \\
    11, 8   &9.87      &33.40   \\
\bottomrule[1.5pt]

\end{tabular}
\caption{Numerical results of SiLLM on De$\rightarrow$En task.}
\label{our_deen}
\end{table}

\begin{table}[]
\centering
\begin{tabular}{p{2cm}<{\centering} p{2cm}<{\centering} p{2cm}<{\centering}} 
\toprule[1.5pt]

\multicolumn{3}{c}{\textit{\textbf{HMT}}}     \\
\hline
    $L, K$   & AL  & SacreBLEU   \\
    1, 2   &0.92      &18.78   \\
    2, 2   &2.42      &22.31  \\
    4, 2    &4.16      &25.29  \\
    5, 4  &5.40      &26.66  \\
    6, 4    &6.29     &27.02  \\
    8, 6   &8.18      &27.80  \\
\midrule[1pt]
\multicolumn{3}{c}{\textit{\textbf{Wait-$k$-SiLLM}}}     \\
\hline
    $k$      & AL  & SacreBLEU   \\
    1      &3.12      &7.00  \\
    3      &4.61      &11.00  \\
    5      &5.96      &12.60  \\
    7      &7.37      &13.00  \\
    9      &8.60      &13.90  \\
\midrule[1pt]
\multicolumn{3}{c}{\textit{\textbf{Wait-$k$-SiLLM+SFT}}}     \\
\hline
    $k$      & AL  & SacreBLEU   \\
    1      &2.47      &4.70  \\
    3      &3.32      &13.70  \\
    5      &4.72      &24.00  \\
    7      &6.30      &26.60  \\
    9      &7.83      &28.10  \\
\midrule[1pt]

\multicolumn{3}{c}{\textit{\textbf{HMT-SiLLM}}}     \\
\hline
    $L, K$    & AL  & SacreBLEU   \\
    1, 2   &2.18      &10.50   \\
    2, 2   &2.64      &11.50  \\
    4, 2   &3.57      &12.20  \\
    5, 4  &4.20      &12.40  \\
    8, 6    &5.98      &13.10  \\
    9, 6   &7.47      &13.60  \\
\midrule[1pt]

\multicolumn{3}{c}{\textit{\textbf{HMT-SiLLM+SFT}}}     \\
\hline
    $L, K$    & AL  & SacreBLEU   \\
    4, 2    &1.58      &22.70  \\
    5, 4   &3.23      &26.50  \\
    6, 4   &4.00      &27.80  \\
    8, 6    &5.70     &29.70  \\
    9, 6   &7.60     &29.70  \\
\bottomrule[1.5pt]

\end{tabular}
\caption{Numerical results of HMT and SiLLM on En$\rightarrow$De task.}
\label{our_ende}
\end{table}

\end{document}